\documentclass[journal]{IEEEtran}
\usepackage{graphicx}
\usepackage{subfigure}
\usepackage{lipsum}
\usepackage{cite}
\usepackage{verbatim}
\usepackage{amsmath}
\usepackage{setspace}
\usepackage{fancyhdr}
\usepackage{hyperref}
\usepackage{color}
\usepackage{amsfonts}
\usepackage{amssymb}
\usepackage{url}
\usepackage{multirow}
\usepackage{graphicx}
\usepackage[normalem]{ulem}

\usepackage[ruled]{algorithm2e} % For algorithms

\SetAlFnt{\small}
\SetAlCapFnt{\small}
\SetAlCapNameFnt{\small}
\SetAlCapHSkip{0pt}
\title{Singularity-Aware Motion Planning for Multi-Axis Additive Manufacturing}

\author{Tianyu Zhang$^{1}$, Xiangjia Chen$^{2}$, Guoxin Fang$^{3,1}$, Yingjun Tian$^{1}$ and Charlie C.L. Wang$^{1\dagger}$,~\IEEEmembership{Senior Member,~IEEE}% <-this % stops a space
\thanks{$^{1}$T. Zhang, G. Fang, Y. Tian and C.C.L. Wang are with the Department of Mechanical, Aerospace and Civil Engineering, University of Manchester, UK}%
\thanks{$^{2}$X. Chen is with the Department of Mechanical and Automation Engineering, Chinese University of Hong Kong, (CUHK) China.}%
\thanks{$^{3}$G. Fang is also a PhD student with the Faculty of Industrial Design, Delft University of Technology, The Netherlands.}%
\thanks{This work was partially supported by HKSAR RGC General Research Fund: 14202219 when the authors worked at CUHK.}
\thanks{$^\dagger$Corresponding Author: {\tt\footnotesize changling.wang@manchester.ac.uk}}
}

\newcommand{\rev}[2]{#2}

\begin{document}
\maketitle
\thispagestyle{plain} \pagestyle{plain} % add page number
%%%%%%%%%%%%%%%%%%%%%%%%%%%%%%%%%%%%%%%%%%%%%%%%%%%%%%%%%%%%%%%%%%%%%%%%%%%%%%%%
\begin{abstract}
Multi-axis additive manufacturing enables high flexibility of material deposition along dynamically varied directions. The Cartesian motion platforms of these machines include three parallel axes and two rotational axes. Singularity on rotational axes is a critical issue to be tackled in motion planning for ensuring high quality of manufacturing results. The highly nonlinear mapping in the singular region can convert a smooth toolpath with uniformly sampled waypoints defined in the model coordinate system into a highly discontinuous motion in the machine coordinate system, which leads to over-extrusion / under-extrusion of materials in filament-based additive manufacturing. \rev{}{The problem is challenging as both the maximal and the minimal speeds at the tip of a printer head must be controlled in motion.} Moreover, collision may occur when sampling-based collision avoidance is employed. In this paper, we present a motion planning method to support the manufacturing realization of designed toolpaths for multi-axis additive manufacturing. Problems of singularity and collision are considered in an integrated manner to improve the motion therefore the quality of fabrication. 
%Experiments are conducted to demonstrate the performance of our method.
\end{abstract}

\section{Introduction}\label{secIntro}

\IEEEPARstart{A}{dditive} manufacturing (AM) has shown significant impact on a variety of industrial applications with its capability in agile fabrication of products with complex geometry~\cite{GIBSON_BOOK14,GAO15_CAD}. The conventional AM setup always conducts three-axis motion and accumulates material in planar layers along the z-axis. Although this \rev{sort of }{}simplification can reduce the cost of hardware system and the complexity of software, it also brings the \rev{problem}{problems} of weak mechanical strength~\cite{zhou2011layerless}, additional supporting structure under overhang~\cite{Hu2015} and stair-case artifacts on the surface~\cite{Etienne19_TOG}. 
%All these problems can be better solved if motions in multiple \textit{degree-of-freedoms} (DOFs) are allowed. 

In recent years, AM systems including multi-axis motion have been developed to overcome the drawbacks of planar layer-based material deposition. Material deposition can be conducted along the normal of curved surface in these systems so that they enable advanced functions such as \rev{}{support-free or supportless printing~\cite{Bhatt_AM20,Dai_ACM18,wu2019general,Li21_CAD,Peng_CHI16,Wang_TVCG18}, strength enhancement~\cite{Zhang_ma20,FANG_SIGG20} and surface quality improvement~\cite{ISA_JMS19,WULLE_CIPR17}.} Different from planar-layer based AM, the \textit{multi-axis additive manufacturing} (MAAM) system not only requires advanced algorithms to generate toolpaths on curved layers but also brings in the challenge of realizing complicated toolpaths on hardware by motion planning. 

\begin{figure}[t]
\centering
\includegraphics[width=\linewidth]{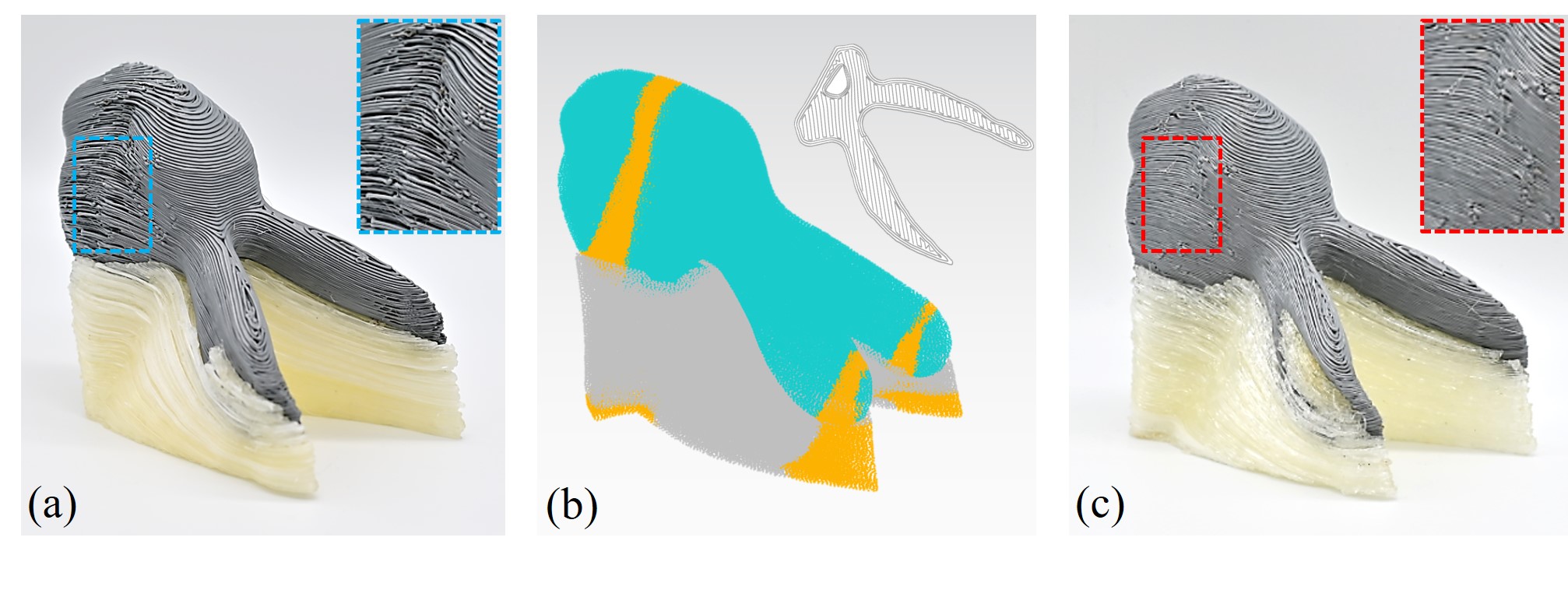}\\
\vspace{-10pt}
\caption{Bunny models fabricated with curved layers by MAAM with rotational tilting table. (a) Surface artifacts (circled region)\rev{, generated by non-smooth material deposition which is sourced to large speed variation of motion. This problem can be solved by applying our singularity-aware motion planning (see the optimized result shown in (b))}{~caused by not well-planed motion. (b) Waypoints in singular region are shown in yellow color. }(c) The surface quality improvement after singularity-aware motion planning.}\label{fig:teaser}
\vspace{8pt}
\end{figure}

\subsection{Problem of Motion Planning in MAAM}\label{subsec:problemDefine}
Without loss of the generality, a toolpath for MAAM can be represented as a set of waypoints containing both position and orientation information represented in the \textit{work-piece coordinate system} (WCS). The waypoints are uniformly sampled on the toolpath to indicate the desired movement of nozzle during the process of material deposition in a speed with less variation. \rev{When the speed variation at nozzle is too large, it is very challenging to synchronize the speed of material extrusion with the speed of nozzle movement.}{}Each waypoint on a toolpath is transferred into the \textit{machine coordinate system} (MCS) by inverse kinematics. However, the mapping between WCS and MCS is nonlinear, especially in the region where the normal direction is close to vertical. This issue \rev{is well-known}{has been studied} in multi-axis \textit{computer numerical control} (CNC) milling as singularity (Sec.~\ref{SingularityIssue}) where solutions has been \rev{welly conducted}{developed in prior research} (see Sec.~\ref{RelatedWork}). However, \rev{those solution}{the existing CNC solutions} cannot be directly applied to MAAM. \rev{}{The reason is threefold.}
\begin{itemize}
    \item \textbf{Continuous Motion:} \rev{}{The filament extrusion during printing process should not be broken; otherwise, both the surface quality and the mechanical strength of 3D printed models will be influenced (ref.~\cite{Silvan18_Nature,FANG_SIGG20}) -- see also the illustration given in Fig.\ref{fig:teaser}.} 
    
    \item \textbf{Speed Constraints:} \rev{}{The feedrate of material deposition can only be stably controlled by the extruder in a range $[f_{\min},f_{\max}]$, which is limited by its working principle in physics. As a consequence, motion speed at the tip of a printer head needs to be well controlled within a corresponding range -- i.e., with both the lower and the upper bounds. To ensure a speed larger than the lower bound becomes challenging in the singular region as it may require a fast rotation that is already beyond the capability of a motor's maximal speed. Details will be discussed in Sec.~\ref{subsecSingularityProcessing}.}
    
    \item \textbf{Collision-free:} \rev{}{Collision avoidance in MAAM is more critical as the working surfaces in freeform shape can be highly complex with many concave regions. When tuning waypoints' normals to optimize motion, collision detection should be systematically integrated in the routine of optimization.}
\end{itemize}
%\rev{}{In the singular region, little variation of orientation in WCS between two neighboring waypoints will result in large rotational movement~\cite{Lin_JAMT14} in \textit{machine coordinate system} (MCS), e.g. C axis. These large movement cannot be achieved in one control step, and middle points are inserted into the neighbor waypoints. Therefore more time is needed to move between neighbor waypoints. In other words, the huge variation of the C axis will force the other axes to move slowly to wait for the C axis in the synchronous motion. As a result, the tangent velocity of contacting point will slow down and extrusion speed limitation is not fulfilled, then stable extrusion cannot be conducted at this low speed. So it will be not helpful to leave alone the singularity problem and just insert more middle points into the toolpath.}

In this work, we present a singularity-aware motion planning pipeline for MAAM, and it is a variant of sampling-based motion planning. Both singularity and collision are considered in an integrated way.\rev{The motion computed by our method satisfies the requirements of 1) collision-free, 2) small variation of speeds on all axes and 3) proper extruder control synchronized with the nozzle's movement.}{} By our approach, \rev{}{the collision-free motion of a given toolpath will be generated to optimize the motion speed at the printer head\footnote{\rev{}{In the singular region, little variation of orientation in WCS between two neighboring waypoints could result in large rotational movement~\cite{Lin_JAMT14} in MCS (e.g., C-axis). This problem of slow motion at the printer head can only be solved by either adjusting the orientations of waypoints (our method) or using a fast enough motor (may cause some problem of dynamic stability). It cannot be solved by adding more waypoints into the toolpath.}} (i.e., falling in the range determined by the feedrates of material extrusion that can be realized). As a result, }the quality of physical fabrication can be significantly improved (see Fig.\ref{fig:teaser} for an example). \rev{}{To the best of our best knowledge, no prior work in CNC literature has directly controlled the minimal speed of motion on a tool.}

\subsection{Related work}\label{RelatedWork}
\rev{}{We review the related work of motion planning on multi-axis machines in both CNC milling and 3D printing.}

\subsubsection{Motion planning in multi-axis CNC}\rev{Each waypoint on a toolpath is transferred into the \textit{machine coordinate system} (MCS) by inverse kinematics. The motion of motors is generated by interpolating these waypoints in MCS. However, the mapping between WCS and MCS is nonlinear in general, which can lead to uneven of axis motion in MCS. For example, singularity happens when the normal of a waypoint is nearly vertical to the working platform. Configurations with nearly vertical orientations in WCS are defined as a singular region, in which even vary small variation of orientation in WCS between two neighboring waypoints will result in large rotational movement in MCS (details are presented in Sec.~\ref{secProblemDefinition}).  As the velocity of the corresponding motor is not fast enough, the planned movement speed at the nozzle cannot be achieved when a toolpath passes through the singular region.}{}For subtractive manufacturing, the singularity and collision issues for multi-axis CNC system have been studied for decades. \rev{For dealing}{To deal} with singularity, singular cone region \rev{is}{was} introduced in~\cite{AFFOUARD_IJMTM04} \rev{where tool-path within this region needs to be modified during motion planning}{}. Sorby et al.~\cite{SORBY_07IJMTM} provided a singularity solution for CNC machine with a non-orthogonal rotary table. 
%\tianyu{These sentences mainly discuss why tool retraction is not suitable and collision should be considered more}
\rev{}{In multi-axis CNC machining, lifting and re-positioning the cutter~\cite{JUNG_02JMPT} is a possible and intuitive solution to solve the singularity issues. However, MAAM has strict requirement on the continuity of motion. This method of retraction cannot be applied here.} 
\rev{}{Boz et al.~\cite{Boz_JAMT13} solved the winding problem and considered the dual IK solutions at each waypoint; however, they did not consider the smoothness of normal variation in singular region. Yang et al.~\cite{YANG_JMTM13} used spline curves in fifth degree to improve the continuous of rotation in the singular region, which however did not utilize the dual IK solutions to decrease the variation of axial motion. Based on the real-time feedback of joint angles, My and Bohez~\cite{CHU_IJPR16} proposed an analytical scheme for identifying and avoiding singular configurations. Grandguillaume et al.~\cite{Grandguillaume_MSF15} solved the singularity problem by controlling the waypoints to going through the singular region while respecting the maximal velocity, acceleration and jerk on the rotary axes. Collision-free is not directly considered in their solution of singularity.}
%\tianyu{show the collision free research}
There are also researches \rev{to obtain}{with focus on generating} a collision-free toolpath. Wang and Tang~\cite{WANG_CAD07} conducted a method to guarantee the collision-free and angular-velocity compliance in the CNC machining process. Potential field is employed in~\cite{LACHARNAY_CAD15} to find a feasible region away from the obstacle when the collision of milling tool is detected. Xu et al.~\cite{XU_IJMS19} proposed a kinematic performance oriented smoothing method to conduct collision avoidance. However, more complex shapes and larger tools make collision detection more complicated in MAAM, and meanwhile collision detection and singularity optimization are coupled together. 

\subsubsection{Motion planning in AM and MAAM}

For the traditional planar-layer based AM, only translation motion is involved in the manufacturing process and it is realized by Cartesian or Delta structures. This machine configuration naturally avoids collision issue and makes the control task of motion easier to solve. \rev{Open-source framework such as Marlin~\cite{Marlin_11MarlinFirmware} already integrates velocity and acceleration optimization function for planar-layer based AM process. In research approaches, advanced motion planning algorithms are conducted for specialized material and fabrication constraints. Constant velocity of nozzle movement is realized in aerosol printing that reduces material waste~\cite{Thompson14_TCPMT}. Restricted continuous deposition of ceramics printing avoiding model collapse is handled in~\cite{Jean19_TOG}. On the other hand, dynamic nozzle extrusion control for planar-layer based AM is studied to ensure the proper material infill on sharp features~\cite{Samuel20_SIG, Kuipers20_CAD}. Advanced planning algorithms are conducted to maximize the continuity of the filament extrusion that can improve the extrusion efficiency and quality~\cite{zhao_sig16}}{Material deposition at sharp features~\cite{Samuel20_SIG, Kuipers20_CAD} and the continuity of filament extrusion~\cite{zhao_sig16} have been well studied for planar-layer based AM}. A more comprehensive survey of motion and toolpath planning in AM can be found in \cite{Jiang20_Micro}. 
%\rev{}{However, these works mainly focus on the planar printing and extrusion volume calculation on each waypoint. Few work takes into account the extrusion speed limitation.} 
When more \rev{}{\textit{degrees of freedom}} (DOFs) are introduced into the material processing, the complexity of motion planning increases sharply for multi-axis machines due to the kinematic redundancy and the collision issues. As a flexible motion platform, robotic arms have been employed to realize multi-axis motion for 3D printing (see~\cite{Bhatt20_ADM} for a survey). Prior works~\cite{XIE_RCIM20, Li21_CAD, Shembekar_CISE19} have provided the smooth path planning and feed-rate control for robot-assisted MAAM system. Huang et al.~\cite{Huang_16TOG} present an optimization-based planning method for robot-assisted frame structure of 3D printing, which finds feasible fabrication sequence to avoid collision. \rev{}{Bhatt et al.~\cite{Bhatt_AM20} adopted a neural network-based scheme to improved both the accuracy and the time lag error for fabricating more accurate parts.} Dai et al.~\cite{Dai20_TASE} developed an algorithm to preserve discrete time constraints when optimizing jerk behavior for the motion of robotic arm. \rev{Differently, our study presented in this paper focuses on parallel multi-axis machines (as shown in Fig.\ref{fig:machnieConfig}) where rotation and translation movements are separated in Cartesian space.}{However, there is less work focusing on motion planning under the speed limitation in singularity region for Cartesian-type multi-axis printers.}  These machine structure (as shown in Fig.\ref{fig:machnieConfig}) are more commonly used for MAAM or hybrid machining as it can generally provide motion with higher precision\rev{, which is better than the hardware setup based on robotic arms}{}.

%\tianyu{Summary the above literature review}
%\rev{}{As discussed above, methods which are feasible for CNC machine, like tool retraction and fixing normal in singular region, cannot get a smooth and continuous motion. Inserting more middle waypoints would slow down the moving speed at the contacting point of material deposition and cannot form a stable extrusion. The other singularity research has rarely demonstrated a unified consideration of the singularity problem and the inherent collision-free requirement. Hence these method developed for subtractive manufacturing cannot be directly applied to MAAM. And to our best knowledge, there are few works to consider collision and singularity region optimization for Cartesian-type MAAM motion planning.}

\subsection{Our approach}\label{subsec:Our approach}
This following contributions are made in this paper:
\begin{itemize}
    \item We present a sampling-based motion planning algorithm to generate a singularity-aware, smooth and collision-free motion by adjusting nozzle orientations of waypoints on a given toolpath, where the coupled problem of collision and singularity are solved systematically.
    
    \item \rev{After building a nozzle extrusion model that considers both the filament volume and the hysteresis property, the requirement of minimal and maximal speed on the nozzle movement is derived to be incorporated in the motion planning.}{
    %Our method optimizes the distribution of extrusion speed between each waypoint, 
    Our algorithm optimizes the motion of machine to satisfy the required range of speeds on the nozzle movement that is derived from the speed limits of material extrusion, and hence improves the surface quality of fabricated models in the singular regions.
    }
\end{itemize}
Our motion planning method is general, which can support MAAM systems in different machine configurations. The effectiveness of our motion planning is demonstrated by the quality of fabrication results and can also be observed from the supplementary video. 

\begin{table*}[t]
    \centering
    \caption{Inverse kinematics of MAAM systems in three different machine configurations}\label{tab:IKSolutions}
    \small
    \begin{tabular}{c|c|c|c}
        \hline
  & Configuration I (Fig.~\ref{fig:machnieConfig}(a)): & Configuration II (Fig.~\ref{fig:machnieConfig}(b)): & Configuration III (Fig.~\ref{fig:machnieConfig}(c)): \\
  &rotational table and tilting  head&rotational and tilting extrusion head&rotational and tilting platform\\
        \hline\hline
        B/C & \multicolumn{3}{c}{$B=  \pm  acos(n_z), \quad \quad C = -atan2({n_y}/{n_x}) \pm \pi H(B)$ } \\
        \cline{2-4}
        X & $p_x \cos C - p_y \sin C + d \sin B$ & $p_x + r \sin C + h \cos C \sin B$ & $ p_x \cos B \cos C -  p_y \cos B \sin C+ p_z \sin B$  \\
        Y & $-p_x \sin C - p_y \cos C $ & $p_y - r \cos C  + h \sin C \sin B + r $ & $ p_x \sin C + p_y \cos C$ \\
        Z & $p_z - d (1-\cos B )$ & $p_z + h \cos B  - h$ & $ p_y \sin B \sin C -p_x \sin B \cos C +p_z \cos B$  \\
        \hline
        \hline
        A/C & \multicolumn{3}{c}{$A=  \pm acos (n_z), \quad \quad  C = -atan2({n_x}/{n_y}) \pm \pi H(A)$ } \\
        \cline{2-4}
        X & $-p_x \cos C + p_y \sin C$ & $p_x - r \cos C + h \sin C \sin A + r$ & $p_x \cos C - p_y \sin C$ \\
        Y & $p_x \sin C +p_y \cos C - d \sin A$ & $p_y - r \sin C  - h \cos C \sin A $ & $p_x \cos A \sin C + p_y \cos A \cos C - p_z \sin A$\\
        Z & $p_z - d (1 - \cos A)$ & $p_z + h \cos A  - h$ & $p_x \sin A \sin C + p_y \sin A \cos C + p_z \cos A $\\
        \hline
    \end{tabular}
\begin{flushleft}\footnotesize
$^{*}$$d$ is the distance between the tip of nozzle and the B-axis, $h$ is the distance between the tip of nozzle and the intersection of B and C axes, and $r$ defines the distance between the tip of nozzle and the C-axis. All these symbols have been illustrated in Fig.~\ref{fig:machnieConfig}.
\end{flushleft}
\vspace{-15pt}
\end{table*}

\section{Motion Planning in MAAM: Problem Analysis}\label{secProblemDefinition}
In the process of MAAM, the nozzle of printer head moves along designed toolpaths to align materials, where each tool-path $\mathcal{L}$ is usually represented by a set of waypoints with both position and orientation information. We remark a single waypoint as {$\mathbf{x} = [\mathbf{p},\mathbf{n}] \in \mathbb{R}^6$} , where $\mathbf{p} = [p_x, p_y, p_z]$ and $ \mathbf{n} = [n_x, n_y, n_z]$ represent the position and the nozzle orientation respectively. \rev{}{Note that $\mathbf{n}$ is a normalized vector in the rest of our paper.}\rev{The speed of nozzle's motion can be estimated by the time used to move from a waypoint $\mathbf{x}_i$ to the next one $\mathbf{x}_{i+1}$, the realization of which on the rotational motors should be well preserved by motion planning.}{} In this section, we first study the problem of motion requirement caused by the control of material extrusion. The  solution of \textit{inverse kinematics} (IK) for three different machine configurations are then presented. Lastly, the coupled issues of singularity and collision are discussed.

\subsection{Extrusion Control and Motion Requirement}\label{subsecExtrusionControl}
For most systems of multi-axis motion, the dynamic control of extra \rev{motion such as extruder}{DOF such as the motor for material extrusion} can be well synchronized with the axial motions. However, the speed of material extrusion is not only limited by the motor of extruder but also many other factors (e.g., the diameter of filaments and the hysteresis property of materials). \rev{}{There is a bounded range of material extrusion speed which could be obtained from experiment as $[f_{min},f_{max}]$.} 
%The requirement on the motion speed of the nozzle is imposed to enable the speed synchronization the axial motion and the material extrusion, which is analyzed below.
%For most multi-axis machine, the dynamic control of extra motion is supported and can be synchronized with axis motion by numerical control system. For 3D printing application, the speed of extrusion is limited thus gives the constrain in axis motion speed. In this section, we present the implement details of control proper extrusion ratio and select motion speed of each axis during synchronized motion.
%Due to the hysteresis and viscosity of materials to be accumulated in MAAM, the range of material \rev{extruder's}{extrusion} speed at the nozzle can be obtained from experiment as $[f_{min},f_{max}]$. 
For MAAM machine, the layer thickness and the toolpath width are dynamically changed. We then estimate the amount of material extrusion between $\mathbf{x}_i$ and $\mathbf{x}_{i+1}$ as 
\begin{equation}
    \Delta E = \frac{k}{4}(T(\mathbf{x}_i)+T(\mathbf{x}_{i+1}) (W(\mathbf{x}_i)+W(\mathbf{x}_{i+1}) \| \mathbf{p}_{i+1} \mathbf{p}_{i} \|
\end{equation}
where \rev{}{$\Delta E$ denotes the volume of extrusion between two waypoints $\mathbf{x}_i$ and $\mathbf{x}_{i+1}$.} $T(\cdot)$ and $W(\cdot)$ are the layer thickness and the toolpath width at a waypoint, and $k$ is a machine-related coefficient that can be obtained by calibration. \rev{Assume the radius of nozzle tip as $R$, we can compute the maximal and the minimal time that are allowed for nozzle motion between two neighboring waypoints as}{The minimally required time and the maximally allowed time to travel between $\mathbf{x}_i$ and $\mathbf{x}_{i+1}$ can be derived as $[t_{\min},t_{\max}] = [{\Delta E}/{f_{\max}},{\Delta E}/{f_{\min}}]$.}

\rev{}{Based on the above analysis, the motion speed $v$ at the tip of a printer head should fall in the following range as $v \in [v_{\min},v_{\max}]$ to achieve a stable material extrusion:}
\begin{equation}
%[t_{min},t_{max}] = [\frac{\Delta E}/{f_{max}},\frac{\Delta E}/{f_{min}}].
\rev{}{[v_{\min},v_{\max}] = \left[\frac{f_{\min} \| \mathbf{p}_{i+1} \mathbf{p}_{i} \|}{\Delta E},\frac{f_{\max} \| \mathbf{p}_{i+1} \mathbf{p}_{i} \|}{\Delta E} \right].}
\end{equation}
The \rev{requirement of $f_{\max}$}{upper bound $v_{\max}$} is imposed to avoid under-extrusion that can be satisfied by \rev{synchronously decreasing the motion speed of nozzle between}{inserting more sample points between} two neighboring waypoints. Moreover, the \rev{requirement of $f_{\min}$}{lower bound $v_{\min}$} is to prevent over-extrusion, which cannot always be achieved when the motion trajectory passing through the singular region. This will be analyzed below and can be achieved by our motion planning algorithm. 

%\rev{}{The time of nozzle's motion between two adjacent waypoints can be determined by the time spent on the axis that takes the most time. Specifically, in the singular region, angle $\Delta C$ is large and always uses most time to move between each waypoints. It should be controlled within $[t_{\min},t_{\max}]$ to follow the requirement of $[f_{min}^*,f_{max}^*]$ and avoid under-extrusion and over-extrusion of materials during deposition.} 

\begin{figure}[t]
\centering
\includegraphics[width=\linewidth]{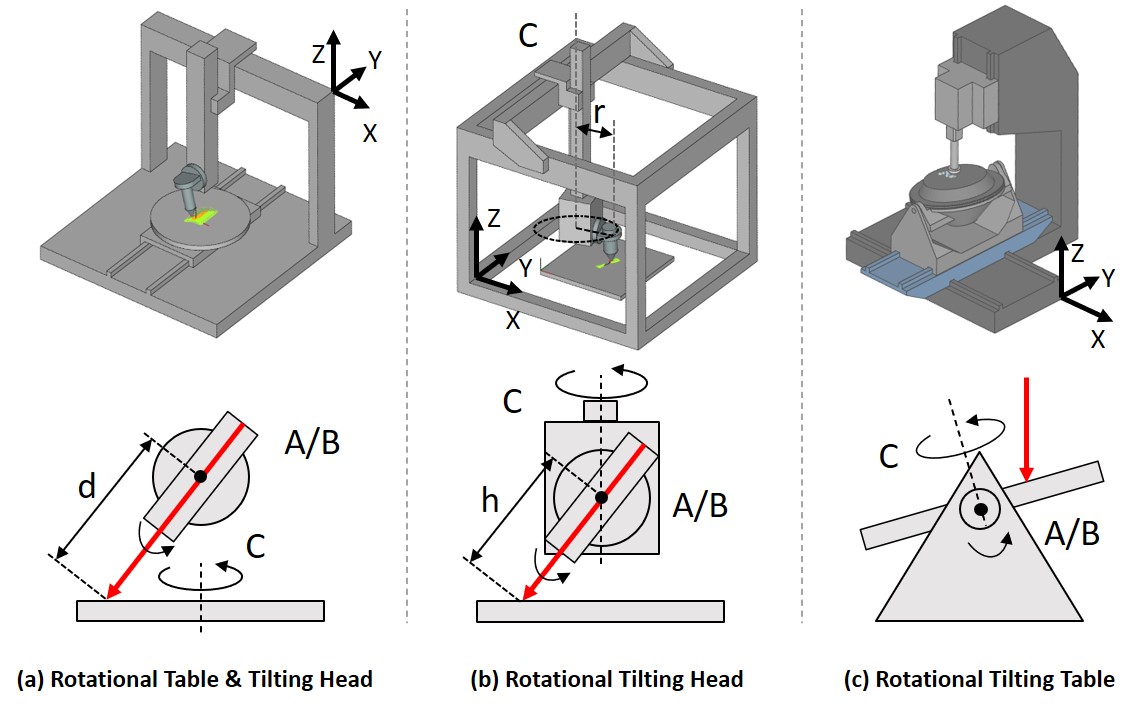}\\
\vspace{-5pt}
\caption{Three different configurations of parallel-based multi-axis motion platforms, where the top row shows their application in MAAM systems. Red arrows in the bottom row are used to denote the nozzle of printer head. }\label{fig:machnieConfig}
\end{figure}

\subsection{Kinematics of Parallel-based Multi-Axis Setups}
The essential step of motion planning is to compute the forward / inverse kinematics of a machine used in MAAM, which actually defines the mapping between \rev{the world coordinate system $\mathbb{R}{^{5}_{WCS}}$}{WCS} and \rev{the machine coordinate system $\mathbb{R}{^{5}_{MCS}}$}{MCS}. In our work, three different parallel-based multi-axis setups are employed to realize linear and rotational movement of the nozzle (see Fig.~\ref{fig:machnieConfig} for the illustration). 

Forward kinematics of machines in these configurations is straightforward. Here we only discuss IK solutions as the singularity problem is caused by it. For a given waypoint $\mathbf{x}$, we can obtain its IK solution on all the three configurations with B and C axes as
\begin{equation}\label{eq:Kinematics}
     B=  \pm \  acos(n_z), \  C = -atan2(\dfrac{n_y}{n_x}) \pm \pi H(B),
\end{equation}
where $H(\cdot)$ is the Heaviside step function. Note that for the configuration with B-axis being replaced by A axial rotation, the IK solution can be obtained by replacing ${n_y}/{n_x}$ with ${n_x}/{n_y}$ in Eq.(\ref{eq:Kinematics}), which gives
\begin{equation}\label{eq:Kinematics2}
     A=  \pm \  acos (n_z) , \  C = -atan2(\dfrac{n_x}{n_y}) \pm \pi H(A).
\end{equation}
The corresponding solutions for linear axis motion (i.e., X, Y and Z) are listed in  %\rev{Table \ref{tab:IKSolutions}}{the supplementary document}.
Table \ref{tab:IKSolutions}.

\subsection{Issue of Singularity}\label{SingularityIssue}
We now analyze the reason of singularity and also the coupled winding issue in the solution of C-axis.

\begin{figure}[t]
\centering
\includegraphics[width=0.65\linewidth]{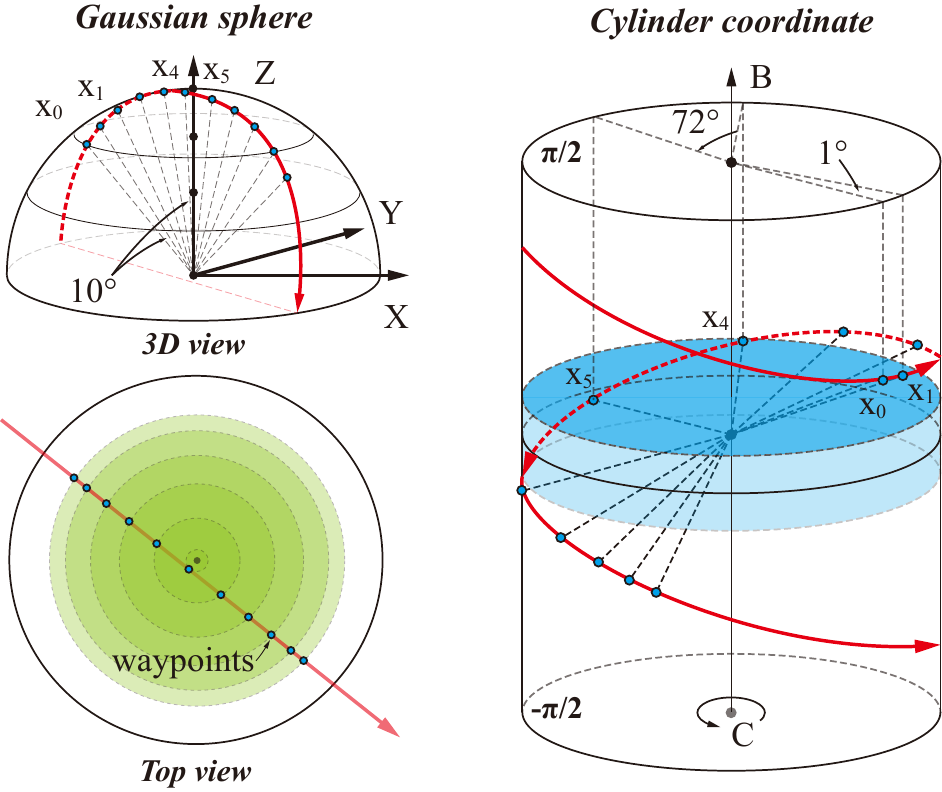}\\
\vspace{-5pt}
\caption{Illustration of the singularity caused by the highly nonlinear mapping of IK \rev{: for a trajectory of uniform variation in orientations (i.e., the orientation difference between $\mathbf{x}_i$ and $\mathbf{x}_{i+1}$ is $10^\circ$), the angle change on C-axis between neighboring waypoints can be highly non-uniform -- e.g., $72^\circ$ between $\mathbf{x}_4$ and $\mathbf{x}_5$ while there is only $1^\circ$ between $\mathbf{x}_0$ and $\mathbf{x}_1$}{solutions}.
}\label{fig:singularityExplanation}
\end{figure}

\begin{figure}[t]
\centering
\includegraphics[width=\linewidth]{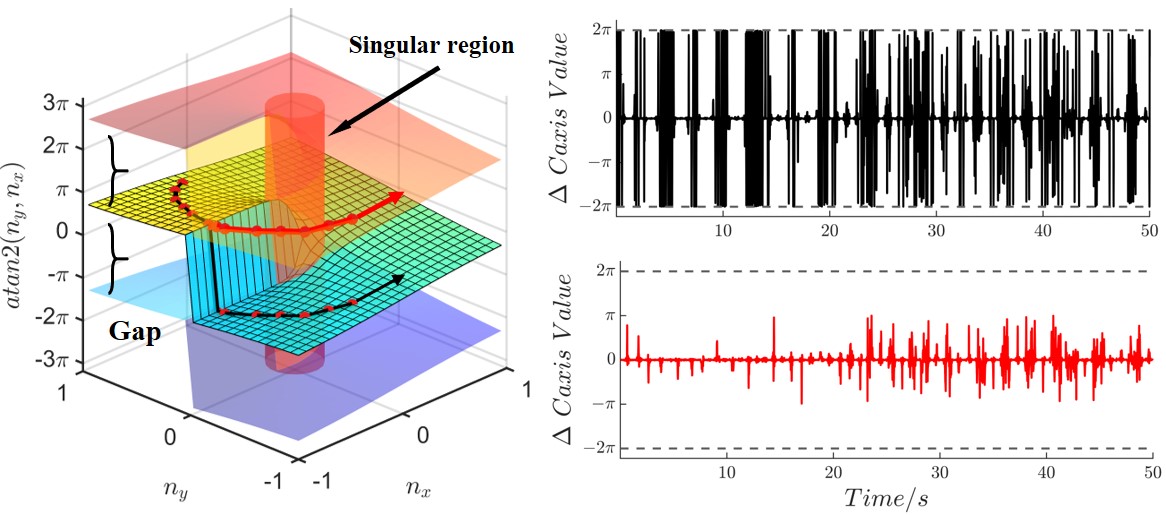}\\
\vspace{-5pt}
\caption{Motion gap of C-axis is caused by the function $atan2()$, and the singular region is visualized as the red cylinder (left). The sharp jumps on C-axis (right-top) can be reduced by considering the winding solution (right-bottom), where the corresponding solutions are also shown in the left as red and black curves.}\label{fig:atan2Func}
\vspace{8pt}
\end{figure}

Directly using the IK solution \rev{provided in Table \ref{tab:IKSolutions}}{(Eqs.\ref{eq:Kinematics} and \ref{eq:Kinematics2})} will result in enormous change of rotational angle when the orientation $\mathbf{n}$ of a waypoint is nearly parallel to Z-axis in WCS. For example as shown in Fig.\ref{fig:singularityExplanation}, the IK solution can map a trajectory with uniform variation in orientations \rev{}{(i.e., $10^\circ$ between any two neighboring waypoints $\mathbf{x}_i$ and $\mathbf{x}_{i+1}$)} into a motion with highly non-uniform angle change on the C-axis\rev{.}{~-- e.g., $72^\circ$ between $\mathbf{x}_4$ and $\mathbf{x}_5$ while there is only $1^\circ$ between $\mathbf{x}_0$ and $\mathbf{x}_1$.} 
This is caused by the nonlinear mapping introduced by the $atan2(\cdot)$ function\rev{around Z-axis in the IK solution}{}. The region of configurations with the angle to Z-axis less than $\alpha$ is defined as the singular region (e.g., the red cylindrical region shown in the left of Fig.~\ref{fig:atan2Func}), where $\alpha$ is a machine-oriented coefficient that can be tuned by experiment \cite{SORBY_07IJMTM}. When the angle change between two neighboring waypoints on C-axis is too large, the motor used for C-axis may not be able to provide the speed \rev{satisfying the condition of synchronized speed on motion and extrusion -- i.e., cannot achieve the requirement of $t_{\max}$ between two waypoints defined in Section \ref{subsecExtrusionControl}. The maximally allowed angle difference on C-axis between two waypoints can be determined by $t_{\max}$ and the maximal speed of the motor for C-axis}{that is fast enough to result in the speed $v > v_{\min}$ at the tip of printer head}. \rev{In our motion planning pipeline}{To solve this problem}, we adjust the orientations of waypoints in the singular region to control the maximal angle difference between neighboring waypoints. 

Considering the property of $atan2(\cdot)$ in Eq.(\ref{eq:Kinematics}), there are discontinuity at $\pm \pi$, which brings sharp jump if $n_x < 0$ and $n_y$ changes its sign between neighbor waypoints (as shown in the left of Fig.~\ref{fig:atan2Func}). Meanwhile, there are multiple solutions of B- and C-axis in Eq.~\ref{eq:Kinematics}. Especially, if the range limitation of C-axis within $[-\pi,+\pi]$ is released, periodic solutions on C-axis can be employed to remove the sharp jump by considering the continuity between configurations of two neighboring waypoints (see the right of Fig.~\ref{fig:atan2Func}). However, winding solutions cannot solve the aforementioned issue of singularity.

In the following section, we present a motion planning method to solve the problem of singularity by adjusting the orientation of waypoints falling in the singular region. The goal is to ensure \rev{}{the moving speed $v$} between two waypoints always be feasible\rev{}{~within the range $[v_{\min},v_{\max}]$}. Moreover, collision should be avoided when changing the orientations of waypoints. Details are \rev{introduced}{discussed} in the following section.

\section{Singularity-aware Motion Planning}\label{secMotionPlanning}
In this section, we present our motion planning method considering both the singularity and the collision issues. A sampling-based strategy is employed here. Every waypoint is converted into one or more points in the machine configuration space \rev{$\mathbb{R}^5_{MCS}$}{MCS}, where some waypoints' orientations will be slightly adjusted. The final trajectory of motion is determined as an optimal path on the graph formed by using these configuration points as nodes, where the total angle variation on B- and C-axes is minimized.

\begin{figure}[t]
    \centering
    \includegraphics[width=0.75\linewidth]{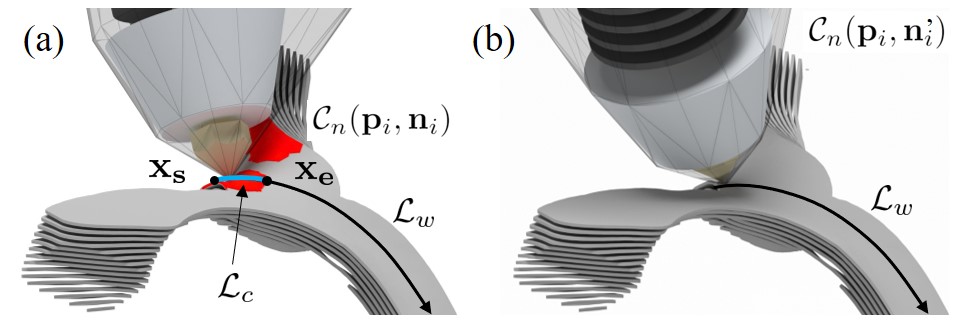}\\
    \vspace{-5pt}
    \caption{\rev{Given waypoints with orientations initially defined by curved-layer slicer as surface normals, collision can happen at those regions with large curvature change (highlighted by red color). We segment each toolpath into collided region $\mathcal{L}_c$ and collision-free region $\mathcal{L}_w$. After applying orientation adjustment, collision-free motion along the whole toolpath can be achieved.}{Eliminating collision by local orientation adjustment. (a) A direction of extruder that will lead to collision (region shown in red). (b) Collision is eliminated after orientation adjustment.}}
    \label{fig:collisionIssue}
    \vspace{8pt}
\end{figure}

\subsection{Collision elimination by orientation adjustment}\label{subsecCollisionElimination}
When using the prior curved-layer slicer~\cite{Dai_ACM18,FANG_SIGG20} to generate the toolpaths for MAAM, the initial orientations of waypoints are computed by the normals of local surface according to the heuristic of applying a locally vertical material adhesion. As only local shape is considered, these initial assignments of waypoints cannot ensure the orientation smoothness throughout the whole toolpath $\mathcal{L}$. Collision also occurs at some local regions (see Fig.~\ref{fig:collisionIssue}(a) for an illustration). Laplacian based smoothing can be applied to the orientations of waypoints to enhance the smoothness of a toolpath but it can also make collision-free waypoints become collided, which will be further processed by the method introduced below.

Collision detection is conducted by modeling a convex-hull of the printer head as $\mathcal{C}$, which is axial symmetry. For a waypoint $\mathbf{x}_i \in \mathcal{L}$, we apply rigid transformation to $\mathcal{C}$ so that the nozzle's tip is located at the $\mathbf{p}_i$ and the rotational axis of $\mathcal{C}$ is aligned with $\mathbf{n}_i$. The transformed convex-hull is denoted by $\mathcal{C}(\mathbf{p}_i,\mathbf{n}_i)$. The collision-indication function $\Gamma(\cdots)$ can be evaluated by  
\begin{align}\label{eq:collectionCheck}
\centering
\Gamma (\mathbf{x}_i)=\left\{ \begin{array}{ll}
0, & \mathcal{C}_n(\mathbf{p}_i,\mathbf{n}_i) \cap \mathcal{M}_i \cap \mathcal{P}  = \emptyset \\
1, & \mathrm{otherwise}
\end{array}\right.,
\end{align}
where $\mathcal{P}$ denotes the platform and $\mathcal{M}_i$ represents the part of model already printed before $\mathbf{x}_i$. 

We can segment the (smoothed) toolpath $\mathcal{L}$ into collided region $\mathcal{L}_c$ and collision-free region $\mathcal{L}_w$. For every waypoint $\mathbf{x}_i=(\mathbf{p}_i,\mathbf{n}_i)$ in $\mathcal{L}_c$, we generate $k$ variants of $\mathbf{x}_i$ as $\tilde{\mathbf{x}}_i^j=(\mathbf{p}_i,\mathbf{n}_i^j)$ with $j=1,\ldots,k$ by randomly sampling $\mathbf{n}_i^j$ on the Gaussian sphere around the point $\mathbf{n}_i$ with $\mathbf{n}_i^j \cdot \mathbf{n}_i \geq \cos \beta$. The value of $\beta$ is used to control the maximally allowed change in orientation (e.g., $\beta=45^\circ$ is employed in our implementation). The collision check is applied to every variant and only the collision-free variant will be kept as candidate waypoints for motion planning. When no collision-free variant is found, we slightly enlarge $\beta$ by 10\% and generate samples again. This step is repeated until a collision-free variant is found. Our motion planning algorithm will select one variant from the set of variants for each waypoint to form the final trajectory while considering the motion smoothness. Details will be presented in Section \ref{subsecMotionPlanning}.

\subsection{Processing waypoints in singular region}\label{subsecSingularityProcessing}
By editing the orientations of the waypoints on a toolpath $\mathcal{L}$, we enhance its smoothness in MCS and also make it collision-free. However, the smoothness of $\mathcal{L}$ in motion (particularly rotational axes) still needs to be further processed due to the singularity issue of $atan2()$ function used in IK. 

Specifically, we define that a waypoint falls in the singular region $\mathcal{R}_s$ if 
\begin{equation}\label{eq:singularPoint}
    \sqrt{\left( {n_x}/{n_z}\right)^2 + \left({n_y}/{n_z}\right)^2} \le tan(\alpha).
\end{equation}
Here $\alpha$ is a small threshold in angle relating to the machine's response speed on C-axis, and $\alpha = 4.5^\circ$ is used in our implementation \rev{by}{according to} experiment. 

We are able to segment a given toolpath $\mathcal{L}$ into the segments inside and out of singular region -- denoted as $\mathcal{L}_s$ and $\mathcal{L} \setminus \mathcal{L}_s$. The last singularity-free waypoint before entering $\mathcal{R}_s$ (denoted by $\mathbf{x}_s$) and the first singularity-free waypoint after leaving $\mathcal{R}_s$ (denoted by $\mathbf{x}_e$) are called \textit{anchor} points as their orientations will \textit{not} be processed. \rev{As shown in Fig.~\ref{fig:singularitySegment},}{} There are three different cases of $\mathcal{L}_s$ including:
\begin{enumerate}
    \item $\mathcal{L}_s$ is connected with waypoints that are out of singular region in both sides,
    
    \item $\mathcal{L}$ ends with $\mathcal{L}_s$, and
    
    \item $\mathcal{L}$ starts from $\mathcal{L}_s$.
\end{enumerate}
The last two are special cases of the first one, where singularity issue can be intuitively solved by aligning the orientation of all waypoints in $\mathcal{L}_s$ \rev{}{to be the} same as the orientation of anchor points $\mathbf{x}_e$ or $\mathbf{x}_s$. In other words, B and C-axis motion are fixed for these special singular regions.

%
%\begin{figure}[t]
%\centering
%\includegraphics[width=\linewidth]{figure/singularitySegment.jpg}
%\caption{Different cases of waypoints along a toolpath falling inside the singular region $\mathcal{R}_s$, where the %segments inside singular region are denoted by $\mathcal{L}_s$.}
%\label{fig:singularitySegment}
%\end{figure}

\begin{figure}[t]
\centering    
\includegraphics[width=0.9\linewidth]{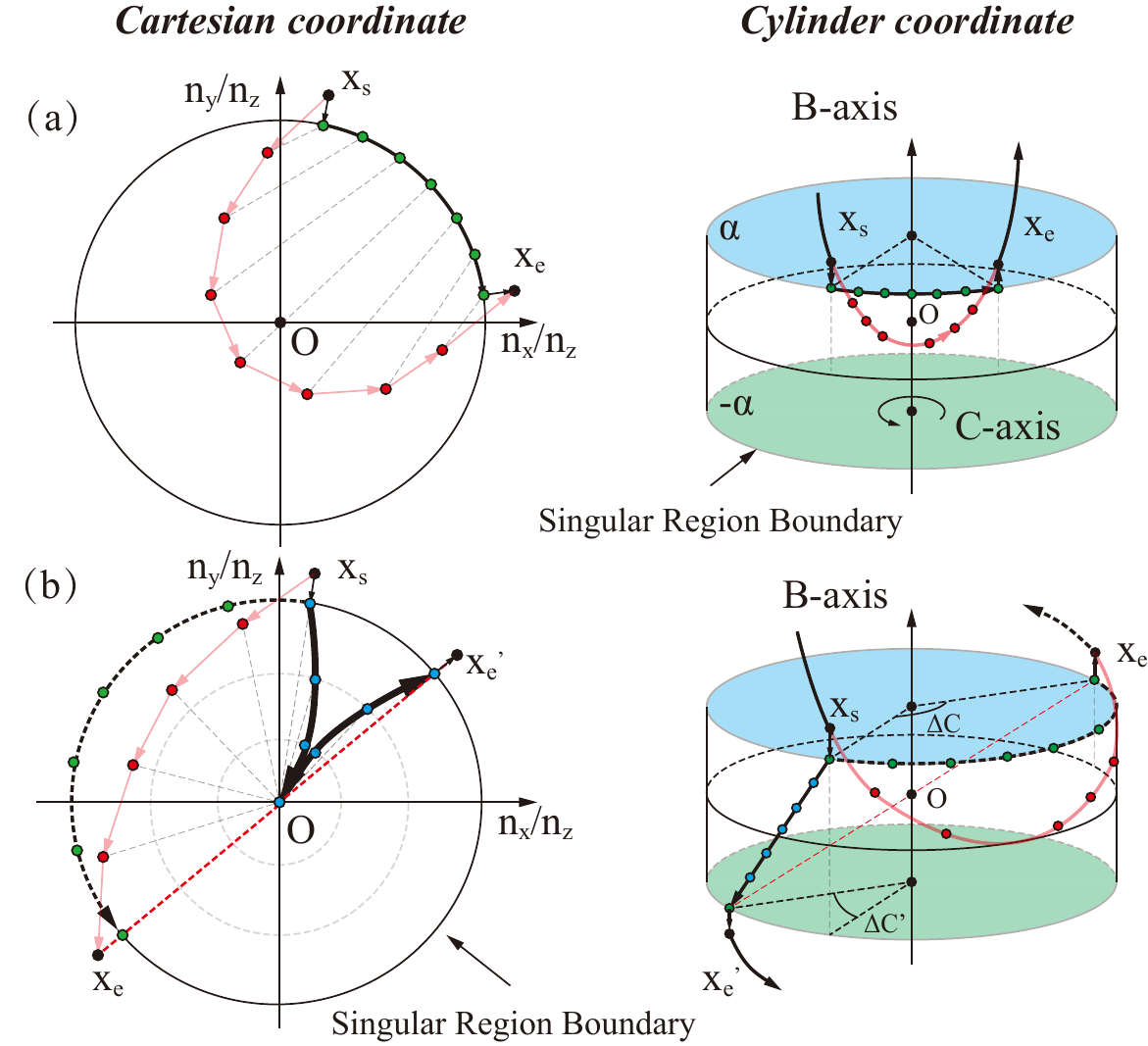}
\caption{Singularity processing for waypoints in $\mathcal{L}_s$. 
(a) Situation with $\Delta C = \|C_s - C_e\| < \frac{\pi}{2}$, \rev{the initial motion (see the polygonal curve shown in red) is improved by}{} pushing the waypoints $\mathbf{x}\in \mathcal{L}_s$ onto the boundary of the singular region\rev{ to form a smooth motion on C-axis (as the bold black curve)}{}. 
(b) By using the other feasible solution (denoted by $\mathbf{x}'_e$) for the anchor point $\mathbf{x}_e$, \rev{ the large variation of angular motion on C-axis can be further reduced from the solution connecting to $\mathbf{x}_e$ that has already been projected onto the boundary of singular region (the dash black curve). In other words, the solution inside singular region connecting to $\mathbf{x}'_e$ but not $\mathbf{x}_e$ is selected as it gives smaller angular variation between neighboring waypoints.}{the waypoints $\mathbf{x}\in \mathcal{L}_s$ go through singular region orderly with less angle variation.}
%The motion represented in Cylindrical coordinate and Cartesian coordinate are shown in the left and right column, respectively.
}\label{fig:singularityProcessing}
\end{figure} 

The singular region defined in Eq.(\ref{eq:singularPoint}) is inside a circle with radius as $\tan \alpha$ in the Cartesian coordinate system (as shown in the left of Fig.~\ref{fig:singularityProcessing}), which can also be represented as a band region as $B \in [-\alpha, +\alpha]$ by the cylindrical coordinates (as shown in the right of Fig.~\ref{fig:singularityProcessing}). For a general case of the toolpath portion $\mathcal{L}_s$ in the singular region with two anchor waypoints $\mathbf{x}_s$ and $\mathbf{x}_e$, we either push every waypoint to the boundary or make another path inside the singular region but with smaller angle change between neighboring waypoints. 

The orientations of processed waypoints in $\mathcal{L}_s$ should smoothly change between $\mathbf{n}_s$ and $\mathbf{n}_e$, which are the orientations of $\mathbf{x}_s$ and $\mathbf{x}_e$ respectively. Assume $C_s$ and $C_e$ are the corresponding IK solution of $\mathbf{x}_s$ and $\mathbf{x}_e$ on C-axis and $\theta(C_e,C_s)$ returns the angle difference between them after considering the winding effect, there are two cases when processing the waypoints in $\mathcal{L}_s$. 
\begin{itemize}
\item $\theta(C_e,C_s) \le \frac{\pi}{2}$: A minor arc is detected by projecting the anchor points, $\mathbf{x}_s$ and $\mathbf{x}_e$, to the boundary of singular region. All waypoints in $\mathcal{L}_s$ are projected onto this arc with equal distance. This ensures that the B-axis motion is locked (i.e., keeping a constant angle $\alpha$) and C-axis has uniform and smooth motion within $\mathcal{L}_s$. An illustration of this projection can be found in Fig.~\ref{fig:singularityProcessing}(a).

\item $\theta(C_e, C_s) > \frac{\pi}{2}$: The above method of projection can also be applied -- see the black dash curve shown in the left of Fig.~\ref{fig:singularityProcessing}(b). However, a solution with smaller angle variation on C-axis can be found by using the other feasible solution of $\mathbf{x}_e$ (denoted by $\mathbf{x}'_e$, which has an inverse value of $B_e$ as $B'_e=-B_e$ and a value of \rev{$C'_e$}{$C_e$} as $C'_e=C_e+\pi$). A newly updated toolpath $\mathcal{L}_s$ can be obtained by generating an interpolation curve between $(B_s,C_s)$ and $(B'_e,C'_e)$ -- see also the illustration in Fig.~\ref{fig:singularityProcessing}(b). This solution is better because of $\theta(C'_e,C_s) < \theta(C_e,C_s)$.
\end{itemize}
After applying this method to process the waypoints in $\mathcal{L}_s$, we can resolve the problem of large angle change on C-axis for most cases. However, there are still some extreme cases having too large angular chance. We break toolpaths in these extreme regions although it rarely happens. 

Changing orientations of the waypoints in $\mathcal{L}_s$ will be possible to generate newly collided configurations. Specifically, if collision occurs at a waypoint $\bar{\mathbf{x}}_i$ processed from $\mathbf{x}_i$, the sampling method introduced in Section \ref{subsecCollisionElimination} is employed to generate $k$ collision-free samples $\{ \tilde{\mathbf{x}}_i^j \}$ varied from $\bar{\mathbf{x}}_i$. To make the newly generated samples closer to $\bar{\mathbf{x}}_i$, the samples are generated within a smaller area (e.g., \rev{$\beta =10^\circ$ is employed here}{with the orientation change less than $10^\circ$}). How to select samples to form the final motion trajectory will be discussed in the next sub-section. 

\begin{algorithm}[t] \label{algorithm:motionPlanning}
\caption{Singularity-aware Motion Planning}
\LinesNumbered
\KwIn{Waypoints for MAAM toolpath $\mathcal{L} = \{ \mathbf{x}_0, \mathbf{x}_1 ... \mathbf{x}_n \}$.}

\KwOut{The best feasible configurations for the waypoints on $\mathcal{L}$ that give a collision-free and smooth motion. 
%Axis control variable [X,Y,Z,B,C] $\forall \mathbf{x}\in \mathcal{L}$ result a collision-free and smooth motion for fabrication. 
}

\tcc{Preprocessing}

Laplacian based smoothing for the orientations of waypoints;

\tcc{Singularity-aware motion processing}

Run singularity check by Eq.(\ref{eq:singularPoint}) and detect the segment $\mathcal L_s$;

Compute initial IK solution for all waypoints in $\mathbb{R}^5_{MCS}$;

\ForEach{$\mathcal{L}_s = \{ \mathbf{x}_s,\mathbf{x}_{s+1},...\mathbf{x}_{e} \}$}{

\eIf{$\mathbf{x}_s = \mathbf{x}_0$ or $\mathbf{x}_e = \mathbf{x}_n$}{
Fix B and C-axis motion with $\mathbf{x}_s$ (or $\mathbf{x}_{e}$).
}{

\tcc{$\mathcal{L}_s$ in-and-out singular zone}

Generating new B- and C-axis coordinates for every waypoints in $\mathcal{L}_s$ by the method in Section \ref{subsecSingularityProcessing}

}

$\forall \mathbf{x}_i \in \mathcal{L}_s$, compute its singular-processed variant $\bar{\mathbf{x}}_i$;
}

\tcc{Generate collision-free variants}

\ForEach{$\mathbf{x}_i \in \mathcal{L}$}{

Run collision check for $\mathbf{x}_i$  by Eq.~\ref{eq:collectionCheck};

\If{$\Gamma (\mathbf{x}_i)>0$}{
Generate $k$ variants for $\mathbf{x}_i$ outside the singular region as $\{\tilde{\mathbf{x}}_i^j\}$;
}

}

\tcc{Graph based search}

Construct $\mathcal{G}$ by waypoints or their collision-free variants;

Apply the Dijkstra's algorithm to compute a shortest path $\mathcal{T}$ on $\mathcal{G}$ which minimizes $J(\mathcal{T})$ defined in Eq.(\ref{eqObjFuncBestTrajectory});

Compute X, Y, Z-axis coordinates for every nodes on $\mathcal{T}$ by IK solution~(Table \ref{tab:IKSolutions});

\Return Optimized [X, Y, Z, B, C] of every nodes on $\mathcal{T}$.

\end{algorithm}

\subsection{Algorithms for Motion Planning}\label{subsecMotionPlanning}
By applying the methods presented in the above two sub-sections, every waypoint $\mathbf{x}_i \in \mathcal{L}$ is in a status as one of the following three cases.
\begin{itemize}
    \item \textit{Case 1}: processed to a variant $\bar{\mathbf{x}}_i$ when $\mathbf{x}_i$ falls in the singular region and $\bar{\mathbf{x}}_i$ is also collision-free;

    \item \textit{Case 2}: processed into $k$ collision-free variants $\{ \tilde{\mathbf{x}}_i^j \}$ if collision occurs at $\mathbf{x}_i$ or collision happens at its variant $\bar{\mathbf{x}}_i$ with singularity processed;
    
    \item \textit{Case 3}: kept unchanged when $\mathbf{x}_i$ is neither in the singular region nor collided.
\end{itemize}
When applying IK computing, every waypoint $\mathbf{x}_i$ (or its $k$ variants) will be converted into $2$ (or $2k$) feasible configurations $\{\mathbf{c}_i^a\}$ in \rev{$\mathbb{R}^5_{MCS}$}{MCS} ($a=1,2$ or $a=1,\cdots,2k$), where each configuration is treated as a sample for motion planning. The final trajectory of motion will be obtained by connecting one selected sample for every waypoints in $\mathcal{L}$. 

A metric is defined as following to evaluate the rotational smoothness of a motion trajectory $\mathcal{T}$
\begin{equation}\label{eqObjFuncBestTrajectory}
    J(\mathcal{T}) = \sum_i |B(\mathbf{c}_i^{T_i}) - B(\mathbf{c}_{i+1}^{T_{i+1}})| + |C(\mathbf{c}_i^{T_i}) - C(\mathbf{c}_{i+1}^{T_{i+1}})| 
\end{equation}
where $T_i$ denotes the index of the selected sample for the trajectory $\mathcal{T}$ at $\mathbf{x}_i$, and the coordinates of rotational axes for a configuration are given by $B(\cdot)$ and $C(\cdot)$. \rev{}{Here we use $L1$-norm instead of $L2$-norm here as $L1$-norm is less sensitive to local errors.} Note that although this metric only evaluates the angular change on B- and C-axes, it also indirectly measures the smoothness of orientation change on the corresponding toolpath which is ensured by the mapping of forward kinematics. Among all possible trajectories, the `best' one gives the smallest value of $J(\cdot)$. 

\begin{figure}[t]
\centering
\includegraphics[width=.95\linewidth]{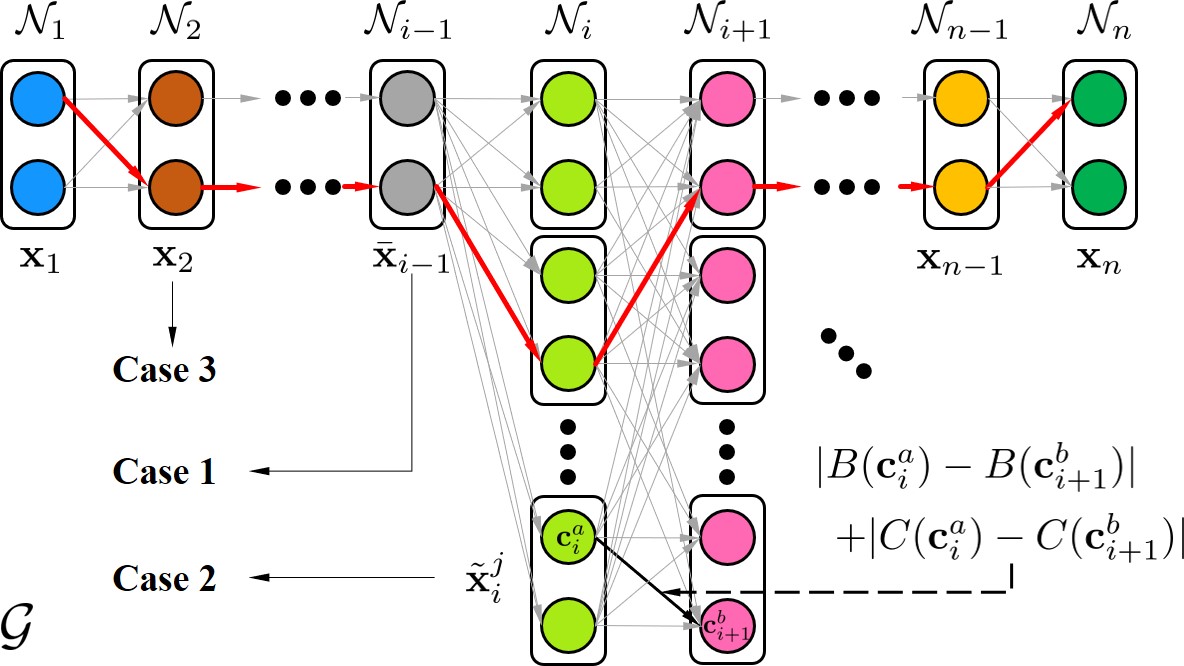}\\
\vspace{-5pt}
\caption{\rev{The best trajectory is searched as a shortest path (displayed in red) on the graph $\mathcal{G}$, each column of which is formed by $2$ (or $2k$) feasible solutions obtained from a waypoint (or its collision-free variants) in the machine configuration space.}{A graph-based algorithm for searching a path (red) on which collision is eliminated and kinematics in singular region is optimized.}
%Candidate normals of one collision waypoint (left) and the red arrow is original collision normal. The Dijkstra-Shortest-Path algorithm in collision-free application (right). \tianyu{The red circle with cross lines means this candidate normal cannot achieve collision-free normal editing, thus it will be discarded.}
}\label{fig:graphForPlanning}
\vspace{5pt}
\end{figure}

A graph-based algorithm is employed to obtain the best trajectory. First of all, the samples of each waypoints in the machine configuration space \rev{$\mathbb{R}^5_{MCS}$}{MCS} are converted into a column of $2$ (or $2k$) nodes on a graph $\mathcal{G}$\rev{}{~as each waypoint has two IK solutions based on  Eqs.(\ref{eq:Kinematics}) and (\ref{eq:Kinematics2})}. The column of nodes for the waypoint $\mathbf{x}_i$ is denoted by $\mathcal{N}_i$. For the toolpath $\mathcal{L}$ with $n$ waypoints, $n$ columns of nodes are constructed (see Fig.\ref{fig:graphForPlanning} for an illustration). Directed edges are added between nodes in neighboring columns. Specifically, for two nodes in two columns as $\mathbf{c}_i^a \in \mathcal{N}_i$ and $\mathbf{c}_{i+1}^{b} \in \mathcal{N}_{i+1}$, a directed edge pointing from $\mathbf{c}_i^a$ to $\mathbf{c}_{i+1}^{b}$ is added into $\mathcal{G}$ with the weight of edge as $|B(\mathbf{c}_i^a) - B(\mathbf{c}_{i+1}^b)| + |C(\mathbf{c}_i^a) - C(\mathbf{c}_{i+1}^b)|$. 

\rev{}{When construct the nodes of $\mathcal{G}$, collision is only considered and prevented at the samples of waypoints and their variants. Although rarely, collision can still occur when there is extremely large change of orientation between two neighboring nodes. To prevent this case, we compute the swept volume of a printer head between two waypoints \cite{Kim04_CAD} while constructing an edge between their corresponding nodes. If collision between this swept volume and the part of model already printed or the platform, we will remove this edge from the graph $\mathcal{G}$. As a result, all candidate paths on $\mathcal{G}$ will be continuous collision-free. In our implementation, we compute the convex hull of printer head in two poses to approximate the general swept volume when the orientation change is small.}

After constructing $\mathcal{G}$ in \rev{this}{the above} way, the optimized trajectory of motion that minimize the objective function $J(\cdot)$ defined in Eq.(\ref{eqObjFuncBestTrajectory}) can be obtained by computing the shortest path on $\mathcal{G}$. The Dijkstra algorithm\rev{}{~\cite{Dijkstra_NM59}} is employed here. The pseudo-code for our singularity-aware motion planning is summarized in \textbf{Algorithm}~\ref{algorithm:motionPlanning}. A collision-free trajectory with smooth motion can be obtained as the output of our method. 

\section{Experimental Results}\label{secResultComparation}
We have implemented the motion planning pipeline for MAAM in C++. Source code of our implementation is released\footnote{\url{https://github.com/zhangty019/MultiAxis_3DP_MotionPlanning}}. Our method can be generally applied to all parallel multi-axis configurations as shown in Fig.~\ref{fig:machnieConfig}, and a simulation platform that can mimic the behavior of multi-axis motion is used to check collision before the physical fabrication (more details can be found in the Supplemental Video). The computation is efficient -- e.g., the motion planning of toolpaths with $41k \sim 434k$ waypoints can be completed in $27.8 \sim 324.9$ sec. on a PC with $2.30$GHz Intel Core i7-10875H CPU and 32GB memory. Experiment of fabrication has been conducted on different models to verify the effectiveness of our approach. 

\begin{figure}[t]
\centering
\includegraphics[width=\linewidth]{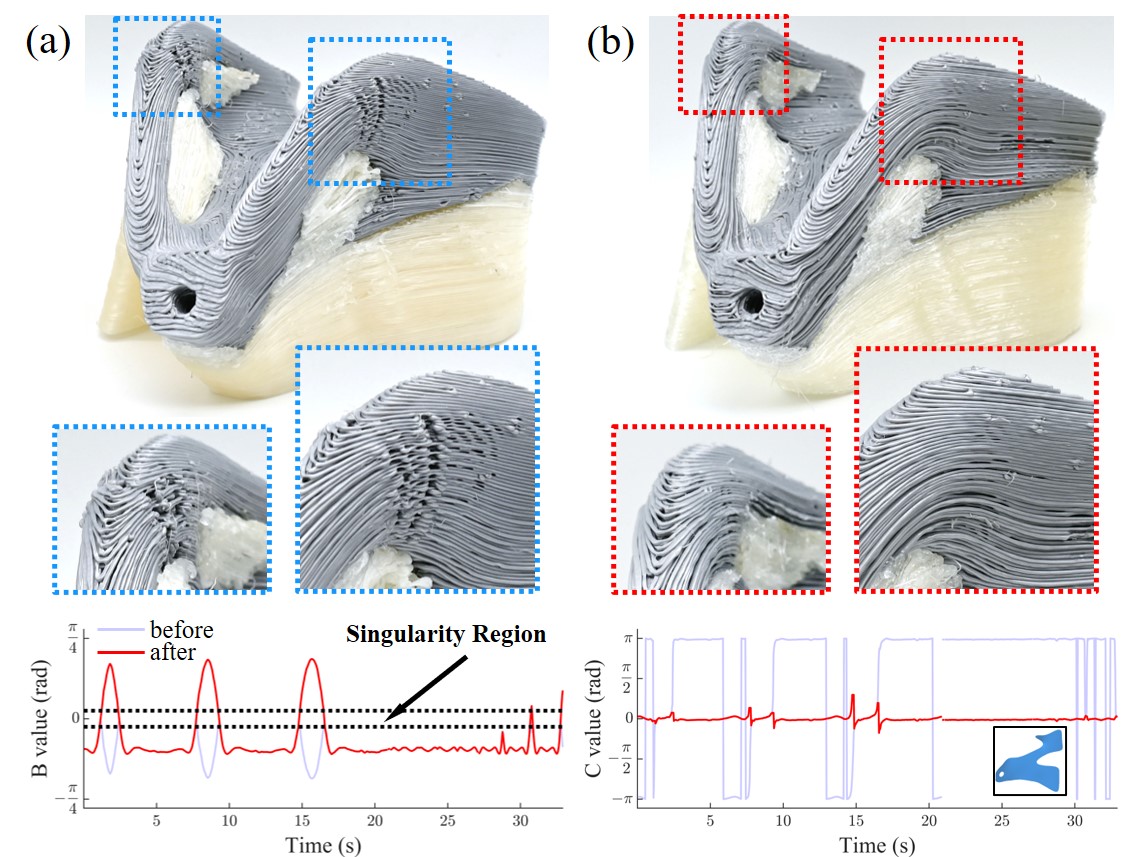}\\
\vspace{-5pt}
\caption{Fabrication result for a mechanical model obtained from topology optimization -- named as topo-opt. Artifacts that damage surface quality and break the continuity of filament can be found in the zoom view of (a), which is significantly \rev{improved}{reduced} by applying our \rev{singularity-aware motion planning method}{method} -- see the result shown in (b). The values of motion on B- and C-axes for a layer of toolpath are compared and given in two graphs at the bottom. \rev{The singular region has been given by dash lines, which is corresponding to the band region of cylinder coordinate shown in Fig.~\ref{fig:singularityProcessing}.}{}}\label{fig:topoModel}
\end{figure}

\begin{figure}[t]
\centering
\vspace{-8pt}
\includegraphics[width=\linewidth]{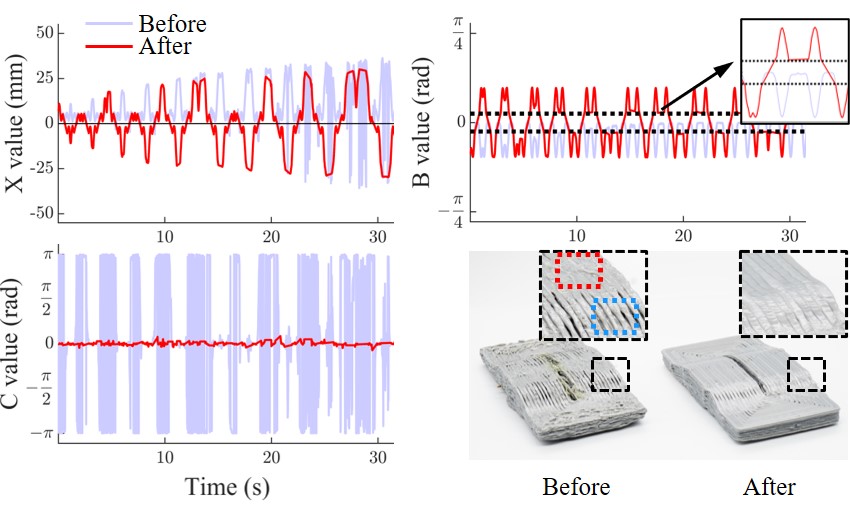}\\
\vspace{-10pt}
\caption{The comparison of motion trajectories generated from orientation smoothed toolpaths, where the curves for the values on X-, B- and C-axes before vs. after applying our method are shown. \rev{There is not too much change on Z-axis thus not shown here.}{} The physically fabricated results shows that both over-extrusion (red rectangular) and under-extrusion (blue rectangular) can be effectively eliminated.}\label{fig:CaxisOptResult}
\vspace{8pt}
\end{figure}

We first compare the results of models fabricated by using trajectories before and after applying our motion planning method. For the trajectory not optimized, it is also processed by Laplacian-based smoothing; but differently, the strategy of \cite{JUNG_02JMPT} is used in singular region by \rev{}{keeping the B- and C-angles unchanged in singular region and }breaking the toolpath between waypoints having large \rev{angle change}{angular variation}. Two models, bunny and topo-opt, are tested and shown in Fig.\ref{fig:teaser} and Fig.\ref{fig:topoModel} respectively. Significant quality improvement can be observed in the regions where toolpath falls into the singular region.

\begin{figure}[t]
\centering
\includegraphics[width=\linewidth]{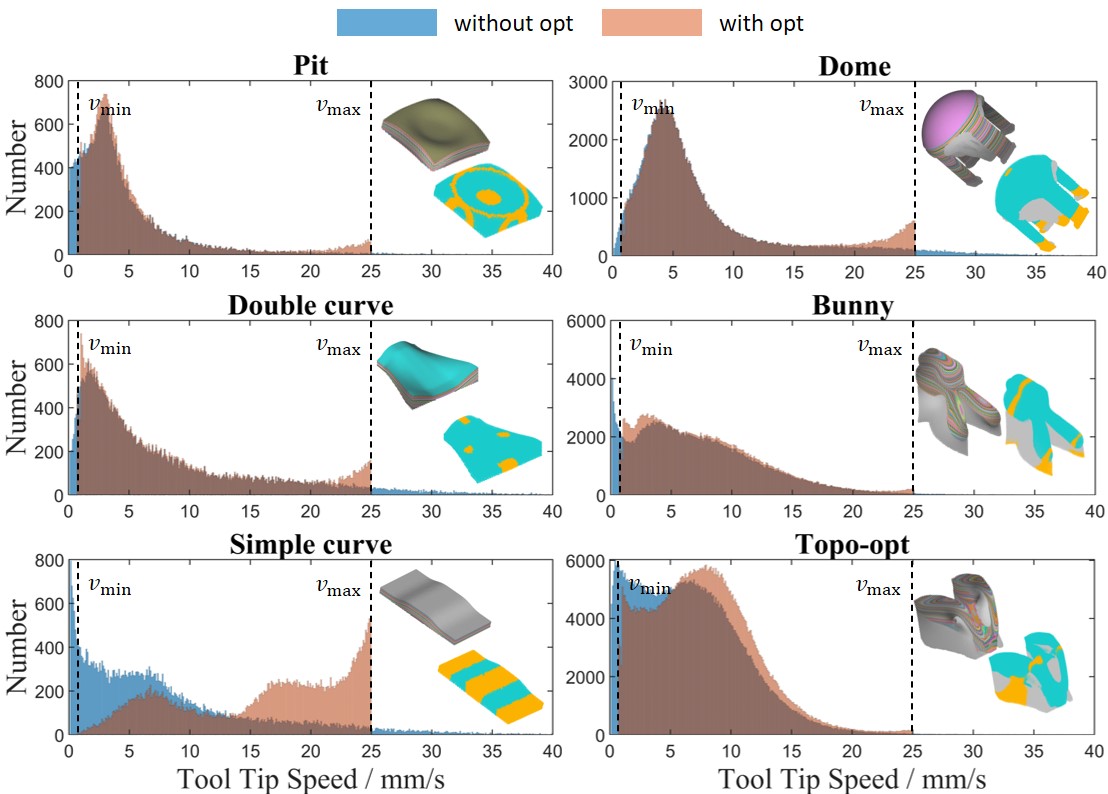}\\
\vspace{8pt}
\footnotesize
\begin{tabular}{ c|c|c||c|c|c } 
 \hline
 \textbf{Model} & without & with opt. & \textbf{Model} & without & with opt. \\ 
 \hline
 {Pit} & 13.07\% & 2.90\%           & {Dome} & 6.90\% & 1.48\% \\ 
 {Double curve} & 11.99\% & 3.05\% & {Bunny} & 10.71\% & 2.03\% \\ 
 {Simple curve} & 17.74\% & 0.14\% & {Topo-opt} & 9.28\% & 1.86\% \\ 
 \hline
\end{tabular}
\caption{\rev{}{Histograms for the speed $v$ of the tip of a printer head at all waypoints with vs. without singularity-aware optimization, where our tests are conducted on six different models with the singular waypoints displayed in yellow. Percentages of waypoints whose speed $v \notin [v_{\min},v_{\max}]$ are reported in the table.}}\label{fig:E_spd_histogram}
\vspace{8pt}
\end{figure}

To quantitatively analyze the behavior of our method in enhancing the smoothness of axial motion, we visualize the values of B- and C-axes before and after applying our singularity-aware motion optimization in Fig.~\ref{fig:topoModel}. \rev{}{The singular region has been given by dash lines, which is corresponding to the band region of cylindrical coordinate shown in Fig.~\ref{fig:singularityProcessing}.} \rev{The C-axis motion}{The motion} has been optimized to \rev{have}{require} much less change \rev{}{on the C-axis} between neighboring waypoints.
%\tianyu{the final angle difference of C angle can be changed, the 15 degree is only a value I give, as the method in Sec III-B can decrease the $\delta$ C into $90^{\circ}$ at the worst case (only one singular point between a pair of anchor points)}
%This not only releases the rotation speed of rational axis in machine but also avoids over-extrusion issue. A close-up look at the behavior of single layer tool-path for Yoga model is shown in Fig.~\ref{fig:layerView}. We remark the region of how different approaches works together in our method to enhance the motion smoothness. \guoxin{more discussion will be added here after Fig.10 is updated.}
The similar analysis is conducted for a model with relatively simpler shape (see the bottom row of Fig.~\ref{fig:CaxisOptResult}), which however has a large area of surface falling in the singular region (i.e., with nearly vertical surface normal). For this example, we do not break a toolpath in singular region even for trajectory directly obtained from IK. Therefore, after generating over-extrusion in single region (see the region circled by red dash lines shown in Fig.~\ref{fig:CaxisOptResult}), it is followed by a portion of under-extrusion that is caused by the hysteresis property of materials (see the region circled by blue dash lines). Both the over-extrusion and the under-extrusion can be eliminated on the result fabricated by using the motion trajectory optimized by our method. Note that the improvement of motion on C-axis is caused by the participate of motion on B-axis; therefore more significant movements occur on B-axis after optimization (see Figs.~\ref{fig:topoModel} and~\ref{fig:CaxisOptResult}). 

\rev{}{We have tested our method on a variety of models. It is found that our approach can effectively change the motion speed at the tip of a printer head to make it within the range of $[v_{\min},v_{\max}]$. As shown in Fig.~\ref{fig:E_spd_histogram}, the percentage of waypoints that violate this speed requirement can be significantly decreased after applying the optimization proposed in this paper. In our tests, $[v_{\min},v_{\max}]=[1.0,25.0] ~(\mathrm{mm/s})$ is employed according to the limited of feedrates that can be provided by the material extruder. Note that, the maximal speed of all motors on X-, Y-, Z-, B- and C-axes  are considered when computing the feasible speed $v$ on a machine here.}
%// X axis 8000 mm/min = 133mm/s 
%// Y axis 6000 mm/min = 100mm/s
%// Z axis 6000 mm/min = 100mm/s
%// B axis 3000 deg/min = 50deg/s
%// C axis 3000 deg/min = 50deg/s

%\charlie{Above edited (May 18)}
%
%\rev{}{Although, the speed limitation of extrusion does not directly affect the solution of motion-planning method, the distribution of nozzle speed in histogram Fig.~\ref{fig:E_spd_histogram} and Table~\ref{tab:percentageOutRange} shows that waypoints concentrate in the range bounded by $ 1.0mm/s$ and $25.0mm/s$ which are the allowable minimum and maximum tip motion speed command respectively. The amount of waypoints with low motion speed where stable extrusion cannot form is decreased greatly. As for the waypoints whose extrusion speed are out of maximum limit can be solved easily by decreasing serve speed of other motion axis.}

%\begin{table}[t]\centering\footnotesize
%\caption{The percentage of tip speed out of limitation}
%\end{table}\label{tab:percentageOutRange}

%Moreover, model fabricated using our result is much stronger since the filament keeps its continuity in critical region. 
%The printing result for yoga model and double-curvature surface model is shown in Fig.~\ref{fig:CaxisOptResult}. It is necessary to mention that these model cannot be fabricated without apply our method as the variation in C-axis rotation brings series collision for high-curvature region.
%\charlie{more discussion to add after applying mechanical tests for the physically printed models.}

\section{Conclusion and Future Work}\label{secConclusion}
To support the manufacturing realization of designed toolpaths for MAAM in different machine configurations, we present a sampling-based motion planning method to solve the problems of singularity and collision in an integrated way. Variants with adjusted orientations are generated for waypoints when needed, and the best trajectory is obtained by connecting the IK solutions with minimal total change of angles on B- and C-axes. As a result, the motion therefore the quality of fabrication can be clearly improved, which has been demonstrated by experimental tests.

We have a few plans to further improve our approach in the future. During the optimization for singularity, the rotation of C-axis is assumed to be unlimited in the current formulation. However, for some machine configuration of MAAM (e.g., Fig.~\ref{fig:machnieConfig}(b)), the motion on C-axis must be constrained due to the twining of electronic cables and material filaments. Constraints for this will be added in our future work. \rev{}{Besides the speed bounds of material extrusion, the acceleration and jerk limitation of actual material extrusion will be considered in our future work of motion planning.}

\bibliographystyle{IEEEtran}
\bibliography{referencePaper.bib}

\begin{thebibliography}{10}
\providecommand{\url}[1]{#1}
\csname url@rmstyle\endcsname
\providecommand{\newblock}{\relax}
\providecommand{\bibinfo}[2]{#2}
\providecommand\BIBentrySTDinterwordspacing{\spaceskip=0pt\relax}
\providecommand\BIBentryALTinterwordstretchfactor{4}
\providecommand\BIBentryALTinterwordspacing{\spaceskip=\fontdimen2\font plus
\BIBentryALTinterwordstretchfactor\fontdimen3\font minus
  \fontdimen4\font\relax}
\providecommand\BIBforeignlanguage[2]{{%
\expandafter\ifx\csname l@#1\endcsname\relax
\typeout{** WARNING: IEEEtran.bst: No hyphenation pattern has been}%
\typeout{** loaded for the language `#1'. Using the pattern for}%
\typeout{** the default language instead.}%
\else
\language=\csname l@#1\endcsname
\fi
#2}}

\bibitem{GIBSON_BOOK14}
I.~Gibson, D.~W. Rosen, and B.~Stucker, \emph{Additive Manufacturing
  Technologies: Rapid Prototyping to Direct Digital Manufacturing},
  1st~ed.\hskip 1em plus 0.5em minus 0.4em\relax Springer Publishing Company,
  Incorporated, 2009.

\bibitem{GAO15_CAD}
W.~Gao, Y.~Zhang, D.~Ramanujan, K.~Ramani, Y.~Chen, C.~B. Williams, C.~C. Wang,
  Y.~C. Shin, S.~Zhang, and P.~D. Zavattieri, ``The status, challenges, and
  future of additive manufacturing in engineering,'' \emph{Comput. Aided Des.},
  vol.~69, pp. 65 -- 89, 2015.

\bibitem{zhou2011layerless}
Y.~Chen, C.~Zhou, and J.~Lao, ``A layerless additive manufacturing process
  based on {CNC} accumulation,'' \emph{Rapid Prototyping Journal}, vol.~17, pp.
  218--227, 2011.

\bibitem{Hu2015}
K.~Hu, S.~Jin, and C.~C.~L. Wang, ``Support slimming for single material based
  additive manufacturing,'' \emph{Comp. Aided Des.}, vol.~65, pp. 1--10, 2015.

\bibitem{Etienne19_TOG}
J.~Etienne, N.~Ray, D.~Panozzo, S.~Hornus, C.~C.~L. Wang, J.~Mart\'{\i}nez,
  S.~McMains, M.~Alexa, B.~Wyvill, and S.~Lefebvre, ``Curvislicer: Slightly
  curved slicing for 3-axis printers,'' \emph{ACM Trans. Graph.}, vol.~38,
  no.~4, July 2019.

\bibitem{Bhatt_AM20}
P.~M. Bhatt, R.~K. Malhan, P.~Rajendran, and S.~K. Gupta, ``Building free-form
  thin shell parts using supportless extrusion-based additive manufacturing,''
  \emph{Additive Manufacturing}, vol.~32, p. 101003, 2020.

\bibitem{Dai_ACM18}
C.~Dai, C.~C.~L. Wang, C.~Wu, S.~Lefebvre, G.~Fang, and Y.-J. Liu,
  ``Support-free volume printing by multi-axis motion,'' \emph{ACM Trans.
  Graph.}, vol.~37, no.~4, July 2018.

\bibitem{wu2019general}
C.~Wu, C.~Dai, G.~Fang, Y.-J. Liu, and C.~C. Wang, ``General support-effective
  decomposition for multi-directional 3-d printing,'' \emph{IEEE Trans. Auto.
  Sci. and Eng.}, vol.~17, pp. 599--610, 2020.

\bibitem{Li21_CAD}
Y.~Li, K.~Tang, D.~He, and X.~Wang, ``Multi-axis support-free printing of
  freeform parts with lattice infill structures,'' \emph{Comput. Aided Des.},
  vol. 133, p. 102986, 2021.

\bibitem{Peng_CHI16}
H.~Peng, R.~Wu, S.~Marschner, and F.~Guimbreti\`{e}re, ``On-the-fly print:
  Incremental printing while modelling,'' in \emph{Proceedings of the 2016 CHI
  Conference on Human Factors in Computing Systems}, 2016, p. 887–896.

\bibitem{Wang_TVCG18}
W.~Wang, Y.-J. Liu, J.~Wu, S.~Tian, C.~C.~L. Wang, L.~Liu, and X.~Liu,
  ``Support-free hollowing,'' \emph{IEEE Trans. Visualization and Computer
  Graphics}, vol.~24, no.~10, pp. 2787--2798, 2018.

\bibitem{Zhang_ma20}
H.~Zhang, D.~Liu, T.~Huang, Q.~Hu, and H.~Lammer, ``Three-dimensional printing
  of continuous flax fiber-reinforced thermoplastic composites by five-axis
  machine,'' \emph{Materials}, vol.~13, no.~7, 2020.

\bibitem{FANG_SIGG20}
G.~Fang, T.~Zhang, S.~Zhong, X.~Chen, Z.~Zhong, and C.~C.~L. Wang, ``Reinforced
  fdm: Multi-axis filament alignment with controlled anisotropic strength,''
  \emph{ACM Trans. Graph.}, vol.~39, no.~6, Nov. 2020.

\bibitem{ISA_JMS19}
M.~A. Isa and I.~Lazoglu, ``Five-axis additive manufacturing of freeform models
  through buildup of transition layers,'' \emph{Int. Journal of Manufacturing
  Systems}, vol.~50, pp. 69 -- 80, 2019.

\bibitem{WULLE_CIPR17}
F.~Wulle, D.~Coupek, F.~Schäffner, A.~Verl, F.~Oberhofer, and T.~Maier,
  ``Workpiece and machine design in additive manufacturing for multi-axis fused
  deposition modeling,'' \emph{Procedia CIRP}, vol.~60, pp. 229--234, 2017.

\bibitem{Silvan18_Nature}
S.~Gantenbein, K.~Masania, W.~Woigk, J.~P. Sesseg, T.~A. Tervoort, and A.~R.
  Studart, ``Three-dimensional printing of hierarchical liquid-crystal-polymer
  structures,'' \emph{Nature}, vol. 561, pp. 226--230, 2018.

\bibitem{Lin_JAMT14}
Z.~Lin, J.~Fu, H.~Shen, and W.~Gan, ``Non-singular tool path planning by
  translating tool orientations in c-space,'' \emph{The International Journal
  of Advanced Manufacturing Technology}, vol.~71, no.~9, pp. 1835--1848, Apr
  2014.

\bibitem{AFFOUARD_IJMTM04}
A.~Affouard, E.~Duc, C.~Lartigue, J.-M. Langeron, and P.~Bourdet, ``Avoiding
  5-axis singularities using tool path deformation,'' \emph{Int. Journal of
  Machine Tools and Manufacture}, vol.~44, no.~4, pp. 415 -- 425, 2004.

\bibitem{SORBY_07IJMTM}
K.~Sørby, ``Inverse kinematics of five-axis machines near singular
  configurations,'' \emph{Int. Journal of Machine Tools and Manufacture},
  vol.~47, no.~2, pp. 299 -- 306, 2007.

\bibitem{JUNG_02JMPT}
Y.~Jung, D.~Lee, J.~Kim, and H.~Mok, ``Nc post-processor for 5-axis milling
  machine of table-rotating/tilting type,'' \emph{Journal of Materials Proc.
  Tech.}, vol. 130-131, pp. 641--646, 2002, aFDM 2002 S.I.

\bibitem{Boz_JAMT13}
Y.~Boz and I.~Lazoglu, ``A postprocessor for table-tilting type five-axis
  machine tool based on generalized kinematics with variable feedrate
  implementation,'' \emph{The International Journal of Advanced Manufacturing
  Technology}, vol.~66, no.~9, pp. 1285--1293, Jun 2013.

\bibitem{YANG_JMTM13}
J.~Yang and Y.~Altintas, ``Generalized kinematics of five-axis serial machines
  with non-singular tool path generation,'' \emph{Int. Journal of Machine Tools
  and Manufacture}, vol.~75, pp. 119--132, 2013.

\bibitem{CHU_IJPR16}
C.~A. My and E.~L. Bohez, ``New algorithm to minimise kinematic tool path
  errors around 5-axis machining singular points,'' \emph{Int. Journal of
  Production Research}, vol.~54, no.~20, pp. 5965--5975, 2016.

\bibitem{Grandguillaume_MSF15}
L.~Grandguillaume, S.~Lavernhe, and C.~Tournier, ``Kinematical smoothing of
  rotary axis near singularity point,'' \emph{Materials Science Forum}, vol.
  836-837, 10 2015.

\bibitem{WANG_CAD07}
N.~Wang and K.~Tang, ``Automatic generation of gouge-free and
  angular-velocity-compliant five-axis toolpath,'' \emph{Comput. Aided Des.},
  vol.~39, no.~10, pp. 841 -- 852, 2007.

\bibitem{LACHARNAY_CAD15}
V.~Lacharnay, S.~Lavernhe, C.~Tournier, and C.~Lartigue, ``A physically-based
  model for global collision avoidance in 5-axis point milling,'' \emph{Comput.
  Aided Des.}, vol.~64, pp. 1 -- 8, 2015.

\bibitem{XU_IJMS19}
J.~Xu, D.~Zhang, and Y.~Sun, ``Kinematics performance oriented smoothing method
  to plan tool orientations for 5-axis ball-end {CNC} machining,'' \emph{Int.
  Journal of Mech. Sci.}, vol. 157-158, pp. 293 -- 303, 2019.

\bibitem{Samuel20_SIG}
S.~Hornus, T.~Kuipers, O.~Devillers, M.~Teillaud, J.~Mart\'{\i}nez, M.~Glisse,
  S.~Lazard, and S.~Lefebvre, ``Variable-width contouring for additive
  manufacturing,'' \emph{ACM Trans. Graph.}, vol.~39, no.~4, July 2020.

\bibitem{Kuipers20_CAD}
T.~Kuipers, E.~L. Doubrovski, J.~Wu, and C.~C.~L. Wang, ``A framework for
  adaptive width control of dense contour-parallel toolpaths in fused
  deposition modeling,'' \emph{Comput. Aided Des.}, vol. 128, p. 102907, 2020.

\bibitem{zhao_sig16}
H.~Zhao, F.~Gu, Q.-X. Huang, J.~Garcia, Y.~Chen, C.~Tu, B.~Benes, H.~Zhang,
  D.~Cohen-Or, and B.~Chen, ``Connected fermat spirals for layered
  fabrication,'' \emph{ACM Trans. Graph.}, vol.~35, no.~4, 2016.

\bibitem{Jiang20_Micro}
J.~Jiang and Y.~Ma, ``Path planning strategies to optimize accuracy, quality,
  build time and material use in additive manufacturing: A review,''
  \emph{Micromachines}, vol.~11, no.~7, 2020.

\bibitem{Bhatt20_ADM}
P.~M. Bhatt, R.~K. Malhan, A.~V. Shembekar, Y.~J. Yoon, and S.~K. Gupta,
  ``Expanding capabilities of additive manufacturing through use of robotics
  technologies: A survey,'' \emph{Additive Manufacturing}, vol.~31, p. 100933,
  2020.

\bibitem{XIE_RCIM20}
F.~Xie, L.~Chen, Z.~Li, and K.~Tang, ``Path smoothing and feed rate planning
  for robotic curved layer additive manufacturing,'' \emph{Robotics and
  Computer-Integrated Manufacturing}, vol.~65, p. 101967, 2020.

\bibitem{Shembekar_CISE19}
A.~V. Shembekar, Y.~J. Yoon, A.~Kanyuck, and S.~K. Gupta, ``{Generating robot
  trajectories for conformal three-dimensional printing using nonplanar
  layers},'' \emph{Journal of Computing and Information Science in
  Engineering}, vol.~19, no.~3, 04 2019.

\bibitem{Huang_16TOG}
Y.~Huang, J.~Zhang, X.~Hu, G.~Song, Z.~Liu, L.~Yu, and L.~Liu, ``Framefab:
  Robotic fabrication of frame shapes,'' \emph{ACM Transactions on Graphics},
  vol.~35, 2016.

\bibitem{Dai20_TASE}
C.~{Dai}, S.~{Lefebvre}, K.~M. {Yu}, J.~M.~P. {Geraedts}, and C.~C.~L. {Wang},
  ``Planning jerk-optimized trajectory with discrete time constraints for
  redundant robots,'' \emph{IEEE Trans. Auto. Sci. and Eng.}, vol.~17, no.~4,
  pp. 1711--1724, 2020.

\bibitem{Kim04_CAD}
Y.~J. Kim, G.~Varadhan, M.~C. Lin, and D.~Manocha, ``Fast swept volume
  approximation of complex polyhedral models,'' \emph{Comput. Aided Des.},
  vol.~36, pp. 1013--1027, 2004.

\bibitem{Dijkstra_NM59}
E.~W. Dijkstra, ``A note on two problems in connexion with graphs,''
  \emph{Numerische Mathematik}, vol.~1, pp. 269--271, 1959.

\end{thebibliography}

\end{document}